\documentclass[a4paper,conference]{IEEEtran}
\usepackage{soul}
\usepackage[utf8]{inputenc}
\usepackage[small]{caption}
\usepackage{booktabs}
\usepackage{algorithm}
\usepackage{algorithmic}
\usepackage{enumitem}
\usepackage{amsmath,amsfonts,amssymb,amsthm}
\usepackage{subcaption}
\DeclareMathOperator*{\argmax}{arg\,max}

\usepackage{nicefrac}
\usepackage{mathtools}
\usepackage{commath}
\usepackage{xcolor}
\usepackage{diagbox}

\usepackage{nameref}
\usepackage{varioref}
\usepackage{hyperref}
\usepackage{cleveref}

\usepackage{makecell}
\definecolor{neworange}{RGB}{253,174,97}
\definecolor{newgreen}{RGB}{26,150,65}
\definecolor{newblue}{RGB}{44,123,182}
\definecolor{newpurple}{RGB}{94,60,153}
\definecolor{newbrown}{RGB}{166,97,26}

\begin{document}
%
\title{JECL: Joint Embedding and Cluster \\ Learning for Image-Text Pairs}

\author{\IEEEauthorblockN{Sean T. Yang}
\IEEEauthorblockA{University of Washington\\
tyyang38@uw.edu}
\and
\IEEEauthorblockN{Kuan-Hao Huang}
\IEEEauthorblockA{University of California, Los Angeles\\
khhuang@cs.ucla.edu}
\and
\IEEEauthorblockN{Bill Howe}
\IEEEauthorblockA{University of Washington\\
billhowe@cs.washington.edu}}


%


\maketitle

\begin{abstract}
We propose JECL, a method for clustering image-caption pairs by training parallel encoders with regularized clustering and alignment objectives, simultaneously learning both representations and cluster assignments. These image-caption pairs arise frequently in high-value applications where structured training data is expensive to produce, but free-text descriptions are common. JECL trains by minimizing the Kullback-Leibler divergence between the distribution of the images and text to that of a combined joint target distribution and optimizing the Jensen-Shannon divergence between the soft cluster assignments of the images and text. Regularizers are also applied to JECL to prevent trivial solutions. Experiments show that JECL outperforms both single-view and multi-view methods on large benchmark image-caption datasets, and is remarkably robust to missing captions and varying data sizes.
\end{abstract}


%
\IEEEpeerreviewmaketitle

\section{Introduction}

We consider multi-modal unsupervised learning for image-text pairs. In many science and engineering applications, images are equipped with free-text descriptions, but structured training labels are difficult to acquire.  For example, the figures in the scientific literature are an important source of information \cite{lee2017phyloparser}, but training labels require specialized knowledge and may evolve frequently.  These figures are, however, equipped with a caption describing the content or purpose of the figure, and these captions can be used as a source of (noisy) supervision.  Other examples include medical imagery (paired with unstructured physician's notes), art (paired with the artist's or curator's description), or archaeological artifacts (paired with researcher's notes).

A simple approach is to ignore the text and cluster the images alone. Unsupervised image clustering has received significant research attention in computer vision \cite{xie2016unsupervised}. However, as we will show, these single-view approaches fail to differentiate semantically different but visually similar subjects on benchmark datasets. On the other hand, using the captions alone (ignoring the image) is rarely considered viable, since captions do not fully describe the content of the image.
Current multi-modal image-text models focus on matching images and corresponding captions for information retrieval tasks \cite{karpathy2015deep,dorfer2018end} rather than unsupervised learning for image-text pairs. Jin et al. \cite{jin2015cross} characterize correlations between image and text using Canonical Correlation Analysis (CCA). However, the textual information for the model comprised semi-structured tags rather than unstructured free-text descriptions.  Free-text descriptions can capture more information to improve a model's performance, but unstructured text is often considered prohibitively noisy in practice: it can contain irrelevant or inconsistent information, and may even be associated with the wrong image.  We find that these challenges have limited the uptake of machine learning in complex, human-intensive domains in science and the humanities. 

We propose \textit{JECL}, a clustering algorithm for image-text pairs that considers both visual features and text features, learning both a vector representation for the pair as well as a clustering. JECL extends prior work on Deep Embedded Clustering (DEC) \cite{xie2016unsupervised}. 
DEC learns a mapping function from the data space to a lower-dimensional feature space and produces soft cluster assignments in which it iteratively optimizes Kullback-Leibler (KL) divergence between soft assignments and computed target distributions. DEC has shown success on clustering several benchmark datasets including both images and text (separately). Despite its utility, we find DEC may often generate empty clusters or singleton clusters containing an obvious outlier, a common problem in clustering tasks \cite{dizaji2017deep}.


JECL learns cluster assignments by iteratively optimizing a clustering objective while learning to align a text distribution~$\mathbf{r}$ and an image distribution $\mathbf{q}$. We address the problem of empty and singleton clusters by introducing regularization terms to force the model to find a solution with a more balanced assignment for each track. We design a target distribution $\mathbf{p}$, such that the model learns by minimizing the KL divergence between $\mathbf{q}$ and $\mathbf{p}$, the KL divergence between $\mathbf{r}$ and $\mathbf{p}$, and the Jensen-Shannon Divergence between $\mathbf{r}$ and $\mathbf{q}$, penalizing the model when these distributions become dissimilar. The final cluster assignments are derived via softmax over the joint distribution (Figure \ref{fig:model}). These combined objectives help JECL define clear boundaries between clusters in the embedding space while retaining semantically meaningful results.  In contrast, DEC can fail to differentiate between visually similar but semantically distinct examples, as we will show.


\noindent We make the following contributions:
\begin{itemize}[leftmargin=1em]

    \setlength\itemsep{0.05em}
    \item We propose JECL, a model that simultaneously learns feature representations and cluster assignments for multi-view data (typically image-text pairs) by penalizing divergence from a shared joint distribution and rewarding coherence across views.
    \item We evaluate JECL on four datasets and compare with multiple single-view and multi-view algorithms, finding that JECL achieves better performance on multiple metrics for all but the smallest dataset due to little data for each cluster.
    \item We show that JECL's performance is robust across many settings, including those with missing text, semantically and visually ambiguous images, sub-optimal hyperparameter settings, and varying data sizes.
\end{itemize}

\section{Related Work}
We consider related work in both image-text representation learning and multi-view clustering methods.


\paragraph{Multi-View Image-Text Representation}
DeVise \cite{frome2013devise} generates visual-semantic embeddings by linearly transforming a visual embedding from a pre-trained deep neural network into the embedding space of the text representation. After DeVise, several visual semantic models have been developed by optimizing bi-directional pairwise ranking loss \cite{kiros2014unifying,wang2016learning} and maximum mean discrepancy loss \cite{tsai2017learning}. Maximizing CCA (Canonical Correlation Analysis) \cite{hardoon2004canonical} is also a common way to acquire cross-modal representation. Yan et al. \cite{yan2015deep} address the problem of matching images and text in a joint latent space learned with deep canonical correlation analysis. Dorfer et al. \cite{dorfer2018end} develop a canonical correlation analysis layer and then apply pairwise ranking loss to learn a common representation of image and text for information retrieval tasks. However, most image-text multi-modal studies focus on matching image and text. Few methods study the problem of unsupervised clustering of image-text pairs. 


Jin et al. considered clustering images by integrating the multimodal feature generation with a Locality Linear Coding (LLC) and co-occurrence association network, multimodal feature fusion with CCA, and accelerated hierarchical k-means clustering~\cite{jin2015cross}. However, the text data they handled are tags instead of the long, noisy, and unreliable free-text descriptions we are interested in. Grechkin et al. proposed EZLearn \cite{grechkin2017ezlearn}, a co-training framework which takes image-text data and an ontology to classify images using labels from the ontology. This model requires prior knowledge of the data in order to derive an ontology.  This prior knowledge is not always available, and can significantly bias the results toward the clusters implied by the ontology. 

\paragraph{Multi-View Clustering} 
JECL can be considered as a form of multi-view clustering, except that multi-view methods often only consider only one data type, typically multiple images of the same object.
Matrix factorization is a common approach to address multi-view clustering problem. Liu et al. \cite{liu2013multi} used a joint matrix factorization with restraints to progressively find the consensus between different views. Zhao et al. \cite{zhao2017multi} developed a deep matrix factorization framework which imposes the non-negative representation of all views to be the same in final layer to maximize the mutual information and is graph regularized to preserve the local geometric structure in each view. BMVC, binary multi-view clustering, is presented by Zhang et al \cite{zhang2018binary} to easily scale to large data by collective encoding views into a compact common binary code space and simultaneously clustering the collaborative binary representations using a matrix factorization model. Spectral clustering has shown significant performance in single view tasks and it also has been explored in multi-view scenario. Brbi\'c et al. \cite{brbic2018} proposes a multi-view spectral clsutering framework that encourages sparsity and low-rank solution and balances the agreement across views. However, most spectral clustering methods are not scalable due to its quadratic complexity. 

JECL is also robust to incomplete multi-view tasks, which have received increasing attention in recent years. DAIMC \cite{hu2018doubly} is a method built on weighted semi-nonnegative matrix factorization. It learns a shared feature matrix for all views and prevents the effect from missing view with $L_{2,1}$-Norm regularized regression. Wang et al. \cite{wang2019spectral} build a bridge between perturbation risk bounds and missing view problems and propose PIC which reduces perturbation risk among all views and learns a consensus Laplacian matrix. 


\section{Method} \label{sec:method}


\begin{figure}[!t]
\centering
    \includegraphics[width=\columnwidth]{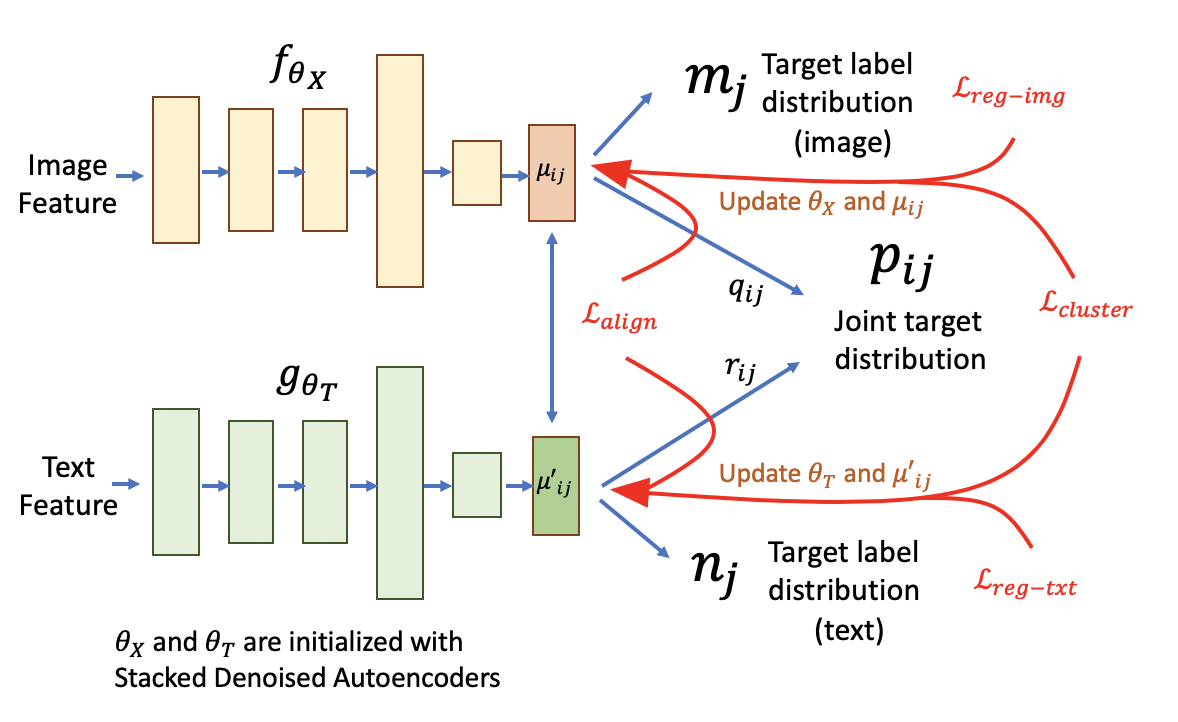}
    \caption{Overview of JECL.  The initialization phase initializes DNN parameters and centroids using a stacked denoising autoencoder and K-means on the embedded data points. During clustering phase, parameters and centroids are updated by minimizing the regularized KL divergence between a joint distribution $\mathbf{p}$ and the image distribution $\mathbf{q}$ (similarly, text distribution $\mathbf{r}$) and the alignment loss between soft cluster assignments of text and images. This figure is best viewed in color.}
    \label{fig:model}%
\end{figure}

Figure \ref{fig:model} shows an overview of our method. JECL clusters data by simultaneously learning 1) DNN parameters \(\theta_X\) and \(\theta_T\) that map each data point with image feature \(x_i\) to an embedding \(z_i \in Z\) and each text feature \(t_i\) to an embedding \(z_i' \in Z'\), and 2) set of image cluster centroids \(\mu_j\)  in \(Z\) and a set of text cluster centroids \(\mu'_j\) in \(Z'\). 



\paragraph{Parameter Initialization}
We initialize DNN parameters \(\theta_X\) and \(\theta_T\) with two stacked denoising autoencoders. Stacked denoising autoencoders have shown success in generating semantically meaningful representations for both text and images in several studies (c.f., \cite{vincent2010stacked,le2013building,xie2016unsupervised}). 
We train the stacked denoising autoencoders to learn the initial DNN parameters for each view by minimizing mean square error reconstruction loss. After training the autoencoders, we discard the decoders, pass data \(x_i\) and \(t_i\) through the trained encoders to obtain the initialized embeddings \(z_i\) and \(z_i'\). Then, we apply K-means to the embeddings  \(z_i\) and \(z_i'\) to obtain initialized centroid sets \(\mu_j\) and \(\mu'_j\). 



\paragraph{Soft Assignment}
Following Xie et al. \cite{xie2016unsupervised}, we model the probability of data point \(i\) being assigned to cluster \(j\) using the Student's t-distribution \cite{maaten2008visualizing}, producing a distribution (\(q_{ij}\) for images and \(r_{ij}\) for text).

\begin{equation}\label{eq:soft_assignment_x}
    q_{ij} = \frac{(1+\norm{z_i-\mu_j}^2/\alpha)^{-\frac{\alpha+1}{2}}}{\sum_{j'}(1+\norm{z_i-\mu_{j'}}^2/\alpha)^\frac{\alpha+1}{2}}
\end{equation}
\begin{equation}\label{eq:soft_assignment_t}
    r_{ij} = \frac{(1+\norm{z'_i-\mu'_j}^2/\alpha)^{-\frac{\alpha+1}{2}}}{\sum_{j'}(1+\norm{z'_i-\mu'_{j'}}^2/\alpha)^\frac{\alpha+1}{2}}
\end{equation}
where \(q_{ij}\) and \(r_{ij}\) are the soft assignments for images and text, respectively, and \(\alpha\) is the number of degrees of freedom of the Student's t-distribution. 

\paragraph{Cluster Alignment}
After calculating the soft assignments for both views, we must align the two sets of $k$ clusters, since the $j$-th image cluster does not necessarily correspond to the $j$-th text cluster. To achieve this, we use the popular Hungarian algorithm \cite{kuhn1955hungarian} to obtain the alignment between image clusters and text clusters. Hungarian algorithm is an optimization algorithm to solve the assignment problem by minimizing the assignment cost. We create a \(k \times k\) confusion matrix where an entry $(m, n)$ represents the number of data points being assigned to $m$-th image cluster and $n$-th text cluster. We then subtract the maximum value of the matrix from the value of each cell to obtain the ``cost.`` The Hungarian algorithm is then applied to the cost matrix to find a clustering assignment with the minimum cost.

\paragraph{Clustering with KL Divergence Minimization}
Similar to Xie et al., we refine the cluster centroids by leveraging high-confidence assignments using an auxiliary joint target distribution.  However, JECL is trained by matching both the image soft assignments $\mathbf{q}$ and the text soft assignments $\mathbf{r}$ to a \emph{single} target distribution $\mathbf{p}$, which allows information passing between the two models.  Specifically, JECL minimizes the KL divergence between $\mathbf{p}$ and $\mathbf{q}$ and the KL divergence between $\mathbf{p}$ and $\mathbf{r}$. The joint loss is as follows:
\begin{align}\label{eq:kl}
    L_{cluster} &= \underbrace{KL(\mathbf{p}||\mathbf{q})}_\text{image loss} + \underbrace{KL(\mathbf{p}||\mathbf{r})}_\text{text loss} \\
    &= \frac{1}{N}\sum^N_i\sum^k_j \Big\{ p_{ij}\log\frac{p_{ij}}{q_{ij}} +  p_{ij}\log\frac{p_{ij}}{r_{ij}} \Big\}
\end{align}



\begin{table*}[h]
    \centering
    \resizebox{0.8\linewidth}{!}{%
    \begin{tabular}{c|c|c|c|c|c}
    Dataset    & \# Points & \# Categories & average \# words & \% of largest Class & \% of smallest Class \\ \hline \hline
    Coco-cross & 7429      & 10     & 50.5 &         23.2 \%             & 1.6\%                \\ \hline
    Coco-all   & 23189     & 43     & 50.4 &        7.4\%               & 0.4\%                \\ \hline
    Pascal     & 1000      & 20     & 48.9 &        5.0\%               & 5.0\%                \\ \hline
    RGB-D      & 1449      & 13     & 38.5 &        26.4\%              & 1.7\%               
    \end{tabular}}
    \caption{Dataset statistics.}
    \label{db-stats}%
    \vspace{-0.5em}
\end{table*}

\paragraph{Choice of Joint Target Distribution}
The target distribution $\mathbf{p}$ aims to improve cluster purity and to emphasize data points with high assignment confidence \cite{xie2016unsupervised}. Our preliminary design adapted this idea to the multi-view problem setting by using a separate target distribution for text and images sub-models separately.  But we found that aligning both images and text to the same joint target distribution simplified the model, improved performance, and was more robust to noise.  The JECL target distribution is as follows.
\begin{equation}\label{eq:joint_target_distribution}
    p_{ij} = \lambda \times \frac{\nicefrac{q^2_{ij}}{\sum_i q_{ij}}}{\sum_{j'} \nicefrac{q^2_{ij'}}{\sum_i q_{ij'}}} + (1 - \lambda) \times\frac{\nicefrac{r^2_{ij}}{\sum_i r_{ij}}}{ \sum_{j'}\nicefrac{r^2_{ij'}}{\sum_i r_{ij'}}}
\end{equation}

where \(\lambda\) is a hyperparameter controlling the relative influence of the images and texts, which we set to 0.5 for all experiments but may be optimized or selected for specialized domains. We found that tuning \(\lambda\) had only a mild effect on performance on our datasets in Section \ref{sec:experimental_results}.  This model is naturally robust to missing text or missing images: Missing text causes the second term in equation (\ref{eq:joint_target_distribution}) to be 0, such that the data points \emph{with} text have a higher value of $p_{ij}$ and therefore contribute a larger gradient to the model. We will show this effect experimentally in Section \ref{sec:experimental_results}. 

\paragraph{Cross-Modality Alignment Loss}

To better exploit the paired features of our data, we apply cross-modality alignment loss to force the soft assignments of every image-caption pair to be similar. The idea is that the text and image from the same pair should be assigned to the same cluster (and, more generally, should have similar soft assignment distributions). We use Jensen–Shannon divergence (JSD) \cite{shannon1998mathematical} to capture the similarity between the cluster assignment distributions:
\begin{align}\label{eq:jsd}
    L_{align} &= JSD(\mathbf{q}||\mathbf{r}) \\
    &= \frac{1}{2}KL(\mathbf{r}||\mathbf{s}) + \frac{1}{2}KL(\mathbf{q}||\mathbf{s})
\end{align}

where $\mathbf{s} = \frac{1}{2}(\mathbf{q} + \mathbf{r})$. JSD is preferred in this setting over KL divergence because it is symmetric and always has a finite value.

\paragraph{Distribution Regularizer}
Many clustering algorithms are prone to producing trivial solutions and empty clusters~\cite{dizaji2017deep,caron2018deep}. Distribution regularizers can significantly improve clustering performance \cite{dizaji2017deep,lin2019bottom} in these situations.
Dizaji et al. used a regularization term to penalize non-uniform cluster assignments. In our case, the overall distribution of the data is unknown and we can not assume that the distribution should be uniform. Instead, we apply a regularizer on each view to avoid empty clusters and maintain freedom for the overall distribution. We define a target label distribution by averaging the soft frequencies for every view. 
\begin{align}\label{eq:label_distribution}
    m_j = q(y = j) &= \frac{1}{N}\sum^N_i r_{ij} \\
    n_j = r(y = j) &= \frac{1}{N}\sum^N_i q_{ij}
\end{align}
where $m_j$ and $n_j$ can be interpreted as the prior frequency of clusters for image and text, respectively. To impose the preference of a balanced assignment, we add a term representing the KL divergence from a uniform distribution \(\mathbf{u}\). The regularized KL divergence is computed as
\begin{align}\label{eq:regularized_kl}
    L_{\text{reg-img}} &= KL(\mathbf{m}||\mathbf{u})  \\
    L_{\text{reg-txt}} &= KL(\mathbf{n}||\mathbf{u}) 
\end{align}

and the overall regularized term can be summarized as:
\begin{equation}\label{eq:reg_loss}
    L_{reg} = L_{reg-img} + L_{reg-txt}
\end{equation}

The overall loss function is as follows.
\begin{equation}\label{eq:overall_loss}
    L_{JECL} = L_{cluster} + \gamma L_{align} + \beta L_{reg}
\end{equation}

where  the first term aims to minimize the dissimilarity between the soft assignment distribution and the joint target distribution as a clustering objective, the second term is to penalize dissimilar soft assignments from the image and text of the same pair, and the last term is to force the model to prefer a balanced assignment for each view to prohibit empty clusters. We have $\gamma$ and $\beta$ as hyperparameters to adjust the weightings of the alignment term and regularized term. We will show that JECL is stable to our hyperparameters in Section \ref{sec:experimental_results}.





\paragraph{Optimization} During the training process, we alternate between two  steps. In the first step, we compute  the target distribution \(p_{ij}\) from \(q_{ij}\) and \(r_{ij}\). In the second step, we fix \(p_{ij}\) to update \(q_{ij}\) and \(r_{ij}\) by refining the parameters (\(\theta_X\) and \(\theta_T\)) and cluster centroids (\(\mu_j\) and \(\mu_j'\)) via gradient descent.  The process continues until convergence.

\paragraph{Final Cluster Assignment} After convergence is met, JECL learns a pair of representations $(z_i, z'_i)$ for each image-text pair $(x_i, t_i)$. The final cluster assignment \(y_i\) can be obtained by 
\begin{equation}\label{label}
    y_i = \argmax_j{p_{ij}}
\end{equation}





\begin{table*}[h]
\centering
{\renewcommand{\arraystretch}{1.1}
\resizebox{0.9\linewidth}{!}{
{\small
\begin{tabular}{cllllllllllll}

\multicolumn{1}{c|}{}  & \multicolumn{3}{c|}{Coco-cross} & \multicolumn{3}{c|}{Coco-all} & \multicolumn{3}{c|}{Pascal} & \multicolumn{3}{c}{RGB-D} \\
\hline
\multicolumn{1}{c|}{}  & \multicolumn{1}{c}{ACC} & \multicolumn{1}{c}{NMI} & \multicolumn{1}{c|}{ARI} & \multicolumn{1}{c}{ACC} & \multicolumn{1}{c}{NMI} & \multicolumn{1}{c|}{ARI} & \multicolumn{1}{c}{ACC} & \multicolumn{1}{c}{NMI} & \multicolumn{1}{c|}{ARI} & \multicolumn{1}{c}{ACC} & \multicolumn{1}{c}{NMI} & \multicolumn{1}{c}{ARI} \\
\hline
\multicolumn{13}{c}{Single-View (Image)}\\
\hline
\multicolumn{1}{l|}{ResNet-50 + KM}  & 0.647  & 0.712    & \multicolumn{1}{l|}{0.558}  & 0.519   & 0.614 & \multicolumn{1}{l|}{0.442} & 0.486 & 0.516 & \multicolumn{1}{l|}{0.307} & 0.353 & 0.289 & 0.161 \\ 
\multicolumn{1}{l|}{ResNet-50 + DEC} & 0.649 & 0.629 & \multicolumn{1}{l|}{0.670} & 0.472 & 0.701 &  \multicolumn{1}{l|}{0.429} & 0.418 & 0.564 & \multicolumn{1}{l|}{0.311} & 0.421 & 0.352 & 0.236     \\
\hline
\multicolumn{13}{c}{Single-View (Text)}\\
\hline
\multicolumn{1}{l|}{Doc2Vec + KM}  & 0.720 & 0.852 & \multicolumn{1}{l|}{0.737} & 0.613 & 0.807 & \multicolumn{1}{l|}{0.589} & \textbf{0.544} & 0.602 & \multicolumn{1}{l|}{0.398} & 0.438 & 0.384 & 0.279  \\
\multicolumn{1}{l|}{Doc2Vec + DEC}  & 0.720 & 0.843 & \multicolumn{1}{l|}{0.729} & 0.557 & 0.738 &\multicolumn{1}{l|}{0.501}  & 0.295 & 0.294 &\multicolumn{1}{l|}{0.120}  & 0.429 & 0.383 & 0.287 \\
\hline
\multicolumn{13}{c}{Concatenation of Both Views + Single-View Models}\\
\hline
\multicolumn{1}{l|}{Concat(ResNet50+Doc2Vec) + KM}  & 0.636 & 0.711 & \multicolumn{1}{l|}{0.550} & 0.517 & 0.617 & \multicolumn{1}{l|}{0.439} & 0.478 & 0.517 & \multicolumn{1}{l|}{0.302} & 0.355 & 0.290 & 0.211  \\
\multicolumn{1}{l|}{Concat(ResNet50+Doc2Vec) + DEC}  & 0.737 & 0.758 & \multicolumn{1}{l|}{0.677} & 0.419 & 0.550 &\multicolumn{1}{l|}{0.275}  & 0.225 & 0.326 &\multicolumn{1}{l|}{0.121}  & 0.344 & 0.255 & 0.172 \\
\hline
\multicolumn{13}{c}{Multi-View Representation Learning} \\
\hline
\multicolumn{1}{l|}{VSE + KM} & 0.665 & 0.736 & \multicolumn{1}{l|}{0.607} & 0.520 & 0.628 & \multicolumn{1}{l|}{0.430} & 0.479 & 0.508 & \multicolumn{1}{l|}{0.300} & 0.388 & 0.318 & 0.194 \\ 
\multicolumn{1}{l|}{DCCA + KM} & 0.712 & 0.822 & \multicolumn{1}{l|}{0.703} & 0.645 & \textbf{0.817} & \multicolumn{1}{l|}{0.603} & 0.442 & 0.485 & \multicolumn{1}{l|}{0.238} & 0.388 & 0.310 & 0.186 \\ 
\multicolumn{1}{l|}{CCAL-\(L_{rank}\) + KM}  & 0.699 & 0.806 & \multicolumn{1}{l|}{0.689} & 0.641 & 0.812 & \multicolumn{1}{l|}{0.587} & 0.446 & 0.489 & \multicolumn{1}{l|}{0.224} & 0.404 & 0.316 & 0.196 \\ 
\hline
\multicolumn{13}{c}{Multi-View Clustering} \\
\hline
\multicolumn{1}{l|}{BMVC}  & 0.365 & 0.227 & \multicolumn{1}{l|}{0.200} & 0.410 & 0.441 & \multicolumn{1}{l|}{0.316} & 0.392 & 0.378 & \multicolumn{1}{l|}{0.214} & 0.207 & 0.088 & 0.047 \\
\multicolumn{1}{l|}{MultiViewLRSSC}  & 0.726 & 0.781 & \multicolumn{1}{l|}{0.706} & 0.569 & 0.747 & \multicolumn{1}{l|}{0.530} & 0.534 & 0.574 & \multicolumn{1}{l|}{0.371} & 0.474 & 0.400 & 0.277 \\
\multicolumn{1}{l|}{DMF-MVC}  & 0.829 & 0.805 & \multicolumn{1}{l|}{0.774} & 0.632 & 0.776 & \multicolumn{1}{l|}{0.608} & 0.512 & 0.573 & \multicolumn{1}{l|}{0.380} & 0.441 & 0.330 & 0.257 \\
\hline
\multicolumn{1}{l|}{JECL}  & \textbf{0.929} & \textbf{0.908} & \multicolumn{1}{l|}{\textbf{0.934}} & \textbf{0.675} & 0.801 & \multicolumn{1}{l|}{\textbf{0.685}} & 0.512 & \textbf{0.625} & \multicolumn{1}{l|}{\textbf{0.403}} & \textbf{0.543} & \textbf{0.472} & \textbf{0.367}  \\ 
\multicolumn{1}{l|}{\quad w/o alignment}  & 0.922 & 0.906 & \multicolumn{1}{l|}{0.931} & 0.634 & 0.784 & \multicolumn{1}{l|}{0.643} & 0.502 & 0.613 & \multicolumn{1}{l|}{0.332} & 0.513 & 0.423 & 0.277  \\ 
\multicolumn{1}{l|}{\quad w/o regularizer}  & 0.894 & 0.890 & \multicolumn{1}{l|}{0.889} & 0.624 & 0.777 & \multicolumn{1}{l|}{0.610} & 0.513 & 0.620 & \multicolumn{1}{l|}{0.376} & 0.520 & 0.433 & 0.327 \\ 
\multicolumn{1}{l|}{\quad w/o alignment \& regularizers} & 0.863 & 0.878 & \multicolumn{1}{l|}{0.852} & 0.611 & 0.757 & \multicolumn{1}{l|}{0.607} & 0.487 & 0.579 & \multicolumn{1}{l|}{0.352} & 0.502 & 0.413 & 0.367 \\ 
\hline 

\end{tabular}
}}}

\caption{Clustering performance of several single-view and multi-view algorithms on four datasets. The results reported are the average of five iterations. JECL outperforms competitive methods on three datasets by a large margin. We also conduct an ablation study on the regularization term and alignment loss. The experimental results show that both additions improve the model significantly.}%
\label{tab:results}%
\vspace{-0.5em}
\end{table*}

\section{Experiments}

We evaluate JECL with four benchmark datasets and compare to both single-view and multi-view algorithms. 

\subsection{Datasets}
To evaluate our method, we choose benchmark datasets that have images with corresponding captions as well as ground-truth labels to define the clusters. We summarize the datasets in Table \ref{db-stats}. (1) \textbf{Coco-cross}: MSCOCO \cite{lin2014microsoft} is a large-scale object detection, segmentation, and captioning dataset. There are five sentences of captions per image. We discard images containing multiple objects and only consider the largest category from ten supercategories. Finally, we have 7,429 data points from these ten categories in total. (2) \textbf{Coco-all}: For this subset of MSCOCO, similar to Coco-cross, we remove images with more than one object, and we remove all categories that include less than 100 images. The result is a dataset with 23,189 images from 43 categories. (3) \textbf{Pascal} \cite{rashtchian2010collecting}: This dataset contains 1,000 images with 20 categories, 50 images each category. Every image is associated with five sentences. (4) \textbf{RGB-D} \cite{KongCVPR14}: This dataset includes 1,449 images with 13 indoor scenes. Every image is captioned with a paragraph which describes the content of the image. Compared to Coco and Pascal datasets, the captions in this dataset are less specific to the categories and significantly less reliable as a source of information. 
\begin{figure*}[h]
\centering
    \includegraphics[width=0.55\linewidth]{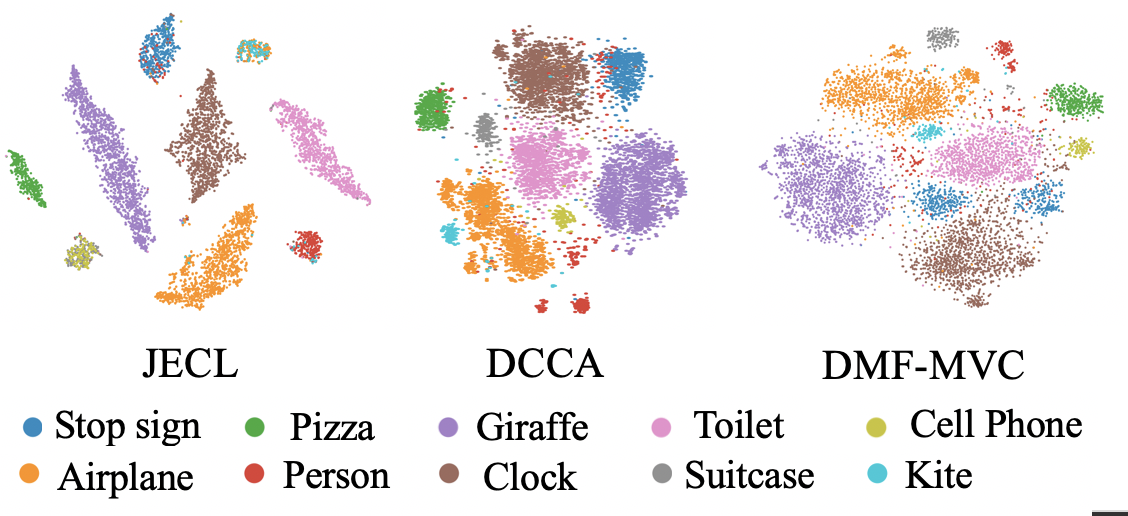}
    \caption{Clustering behavior of JECL, DCCA and DMF-MVC. Color indicates ground-truth labels. Cluster shape and position is not meaningful. JECL successfully separates semantically distinct clusters with clear boundaries between clusters.  While DCCA and DMF-MVC are able to gather semantic similar images, the boundaries between clusters are unclear, which is reflected in the quantitative performance. }
    \label{fig:embeddings}%
    \vspace{-0.7em}
\end{figure*}
\subsection{Competitive Methods}
We compare our method to a variety of  single-view and multi-view methods.
\paragraph{Single-View Methods}
We run two single-view methods to serve as baseline comparisons: \textbf{K-means} (KM) \cite{lloyd1982least} and \textbf{Deep Embedded Clustering} (DEC) \cite{xie2016unsupervised}. 


\paragraph{Multi-View Methods}
We evaluate six state-of-the-art multi-view methods, including three multi-view representation learning models and three multi-view clustering models.
We also evaluate a naive baseline for multi-view methods that simply concatenates the ResNet-50 features and the Doc2Vec features before applying K-means and DEC as in the single-view case. (1) \textbf{VSE} \cite{kiros2014unifying}: Unifying Visual-Semantic Embeddings unifies joint image-text embedding models by minimizing pairwise ranking loss. K-means is implemented to acquire the cluster centroids. (2) \textbf{DCCA} \cite{andrew2013deep}: Deep Canonical Correlation Analysis learns complex nonlinear transformations of two views of data by maximizing the regularized total correlation. The cluster assignments are obtained with K-means on the joint representations. (3) \textbf{CCAL-\(L_{rank}\)}: This method learns a joint representation by maximizing Canonical Correlation with a pairwise ranking loss. K-means is applied to the learned embeddings. (4) \textbf{DMF-MVC} \cite{zhao2017multi}: Multi-View Clustering via Deep Matrix Factorization is a deep matrix factorization framework to learn the semantic information of all views in a layer-wise fashion. (5) \textbf{BMVC} \cite{zhang2018binary}: Binary Multi-view Clustering is a joint learning framework simultaneously compressing inputs into collaborative binary representations and clustering the collaborative representations using a binary matrix factorization model. (6) \textbf{MLRSSC} \cite{brbic2018}: Multi-view Low-rank Sparse Subspace Clustering is a multi-view spectral clustering framework that learns a joint subspace representation by building affinity matrix among all views with the constraints of low-rank and sparsity.

\subsection{Evaluation Metrics}
All experiments are evaluated by three standard clustering metrics: clustering accuracy (ACC), normalized mutual information (NMI), and adjusted rand index (ARI). For all metrics, higher numbers indicates better performance.

\begin{figure*}[t]
\centering
    \includegraphics[width=.45\linewidth]{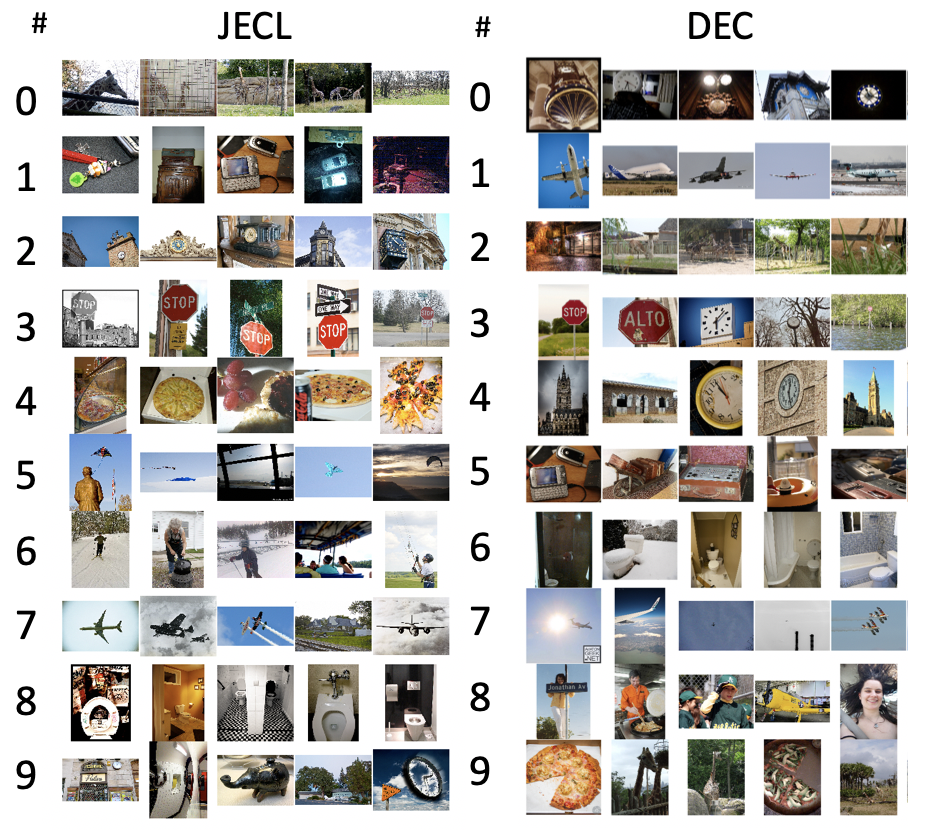}
    \caption{The 5 highest-confidence images in each cluster from JECL and DEC. JECL clusters appear qualitatively better. For example, airplanes and kites, two visually and semantically similar concepts, are clearly distinguished, while DEC appears to struggle to distinguish giraffes and pizza.}
    \label{fig:JECLvsDEC}%
    \vspace{-0.7em}
\end{figure*}
\begin{figure}[h]
\centering
\includegraphics[width=0.8\columnwidth]{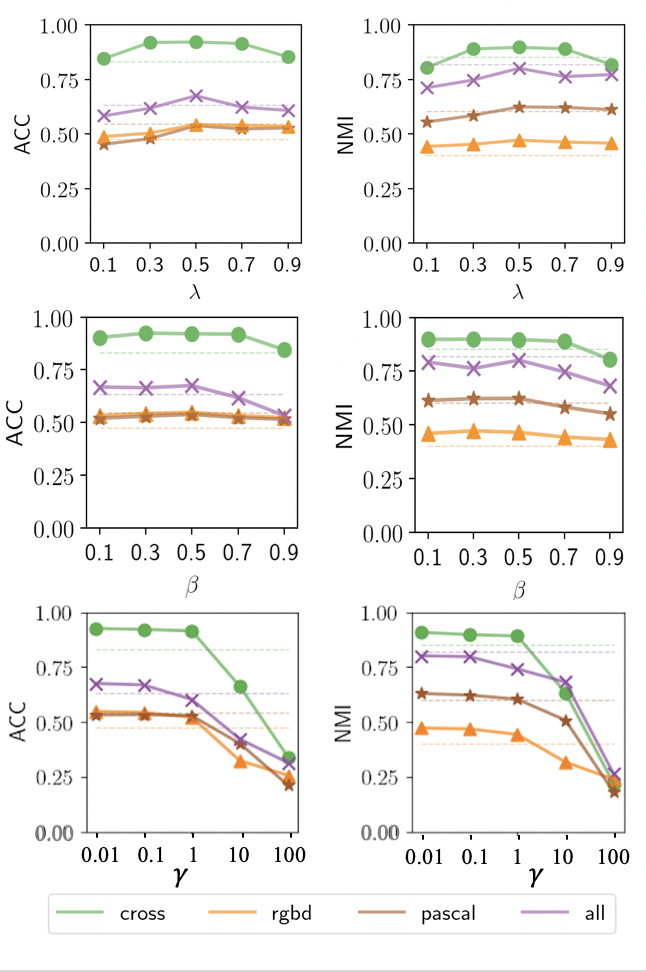}
\caption{The experimental results of hyperparameter sensitivity. The dash lines are the best performing competitive algorithms listed in Table \ref{tab:results}. JECL is generally robust to hyperparameter settings, while is the most stable and produces top results with $\lambda = 0.5$, $\beta = 0.1$ and $\gamma = 0.1$ among all datasets. }
\label{fig:pars}%
\vspace{-0.5em}
\end{figure}

\begin{figure}[t]
\centering
  \begin{subfigure}[t]{0.4\columnwidth}
    \includegraphics[width=\columnwidth]{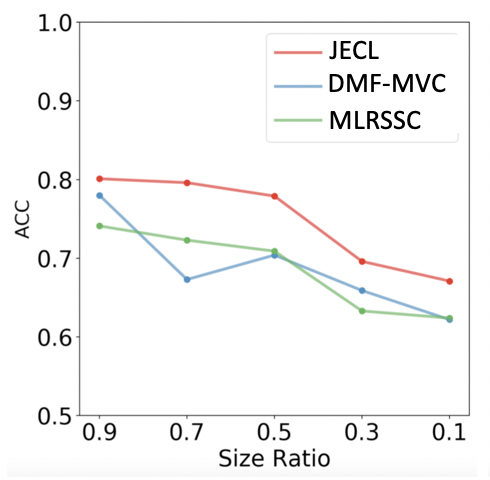}
    \caption{}
    \label{fig:size_ACC}%
  \end{subfigure}
  \hspace{1em}
  \begin{subfigure}[t]{.4\columnwidth}
    \includegraphics[width=\columnwidth]{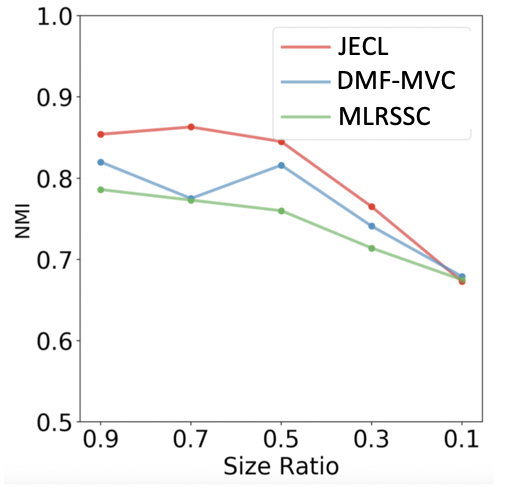}
    \caption{}
    \label{fig:size_NMI}%
  \end{subfigure}
  
\caption{JECL performance as data size decreases. The performance degrades when size ratio is below 0.5 (500 data points in each class), while JECL still outperforms the state-of-the-art multi-view clustering methods, DMF-MVC and MLRSSC on varying data sizes. }
\label{fig:size}%
\vspace{-0.5em}
\end{figure}

\begin{figure}[hbt]
        \centering
        \includegraphics[width=.8\columnwidth]{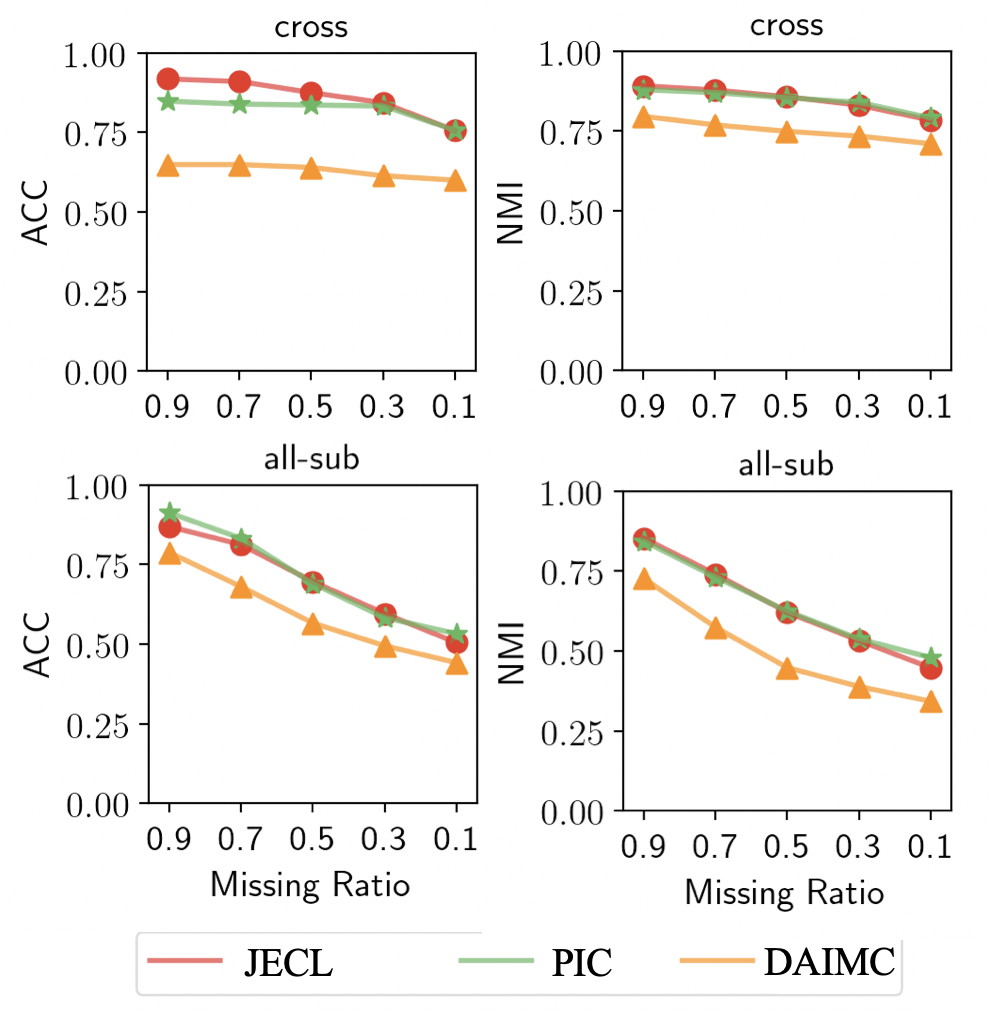}
        \caption{Experimental results on missing view scenarios. JECL is competitive with the state-of-the-art method, PIC, and outperforms DAIMC by a large margin on both datasets.}
        \label{fig:missing}%
        \vspace{-0.7em}
\end{figure}

\begin{figure}[hbt]
        \centering
        \includegraphics[width=.5\columnwidth]{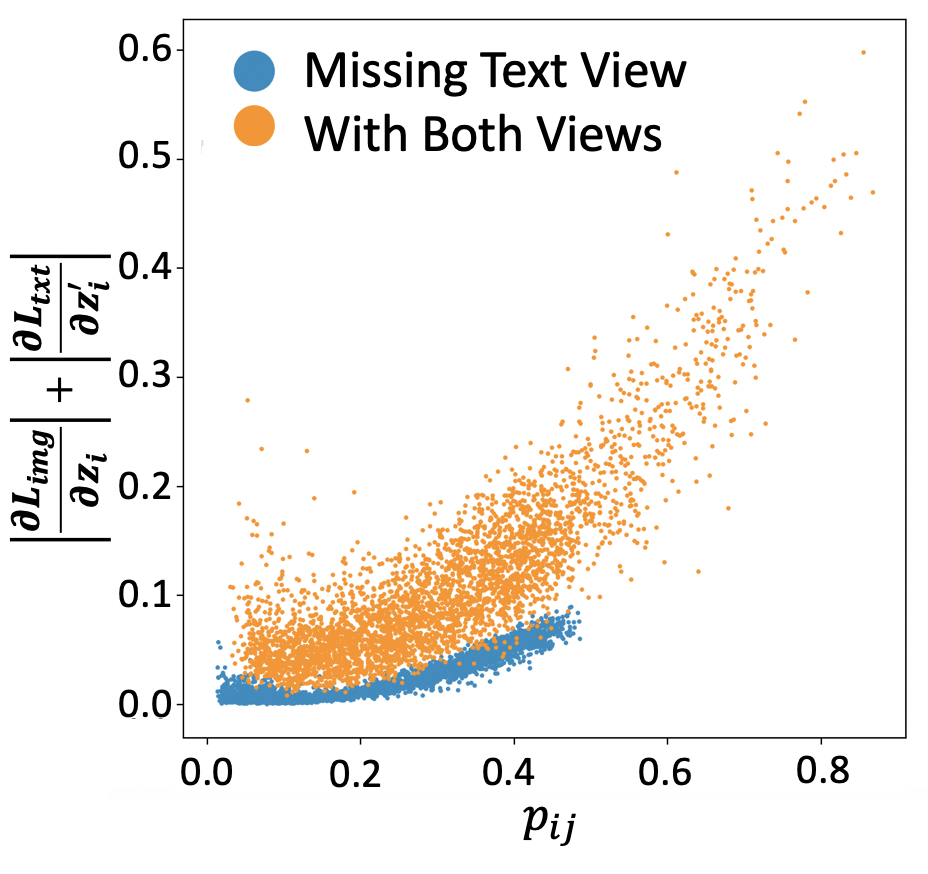}
        
    \caption{JECL's robustness to missing data is attributable to the model of the joint distribution: the images with text (orange) contribute more to the gradient than the images with missing text (blue).}
    \label{fig:missingGradients}%
    \vspace{-0.7em}
\end{figure}

\subsection{Implementation}
We use a ResNet-50 model \cite{he2016deep}, pretrained on the 1.2M images from ImageNet \cite{deng2009imagenet}, to extract 2048-dimensional images features and Doc2vec \cite{le2014distributed}, pretrained on Wikipedia via skip-gram, to obtain 300-dimensional text features. All methods are fed with these pre-trained features except for VSE, which has raw text as input. We use hyperparameter settings following Xie et al. \cite{xie2016unsupervised} for DEC components. \(\lambda\), $\gamma$, and $\beta$ are set to 0.5, 0.1, and 0.1 in all experiments, respectively. We will show that the model remains stable to different hyperparameter settings within reasonable range in next section. For baseline algorithms, we use the same setting in their corresponding paper. All the results are the average of 5 trials . 


\subsection{Experimental Results} \label{sec:experimental_results}

Table \ref{tab:results} displays the quantitative results for different methods on various datasets. JECL outperforms other tools on almost every dataset by a significant margin. The table also contains the results of an ablation study of the distribution regularizer and distribution alignment loss. We can see that both additions improve the overall performance.


\paragraph{Qualitative Comparison} The cluster metrics are difficult to interpret, so we are interested in exploring a qualitative comparison between JECL and the best single-view (DEC), image-text representation learning (DCCA), and multi-view (DMF-MVC) competitors. Figure \ref{fig:embeddings} is a visualization of the latent space of JECL to illustrate its effectiveness in producing coherent clusters. We use t-SNE to visualize the embeddings from the latent space. The positions and shapes of the clusters are not meaningful due to the operation of t-SNE. JECL is able to generate semantically distinct clusters with clear boundaries between clusters. While DCCA and DMF-MVC are able to associate semantically similar images, the cluster boundaries are less distinct. We further compare JECL to DEC to examine the effect of additional view and our alignment loss by inspecting examples of the clusters.  Figure \ref{fig:JECLvsDEC} shows the top five images with highest confidence from each cluster from the \textit{Coco-cross} dataset. The figure shows that DEC clusters are not always coherent. For example, cluster \#1 and cluster \#7 seem to include mostly \textit{airplane} images and cluster \#0 and cluster \#4 are \textit{clock} clusters. Cluster \#9 from DEC is a fusion of \textit{giraffe} and \textit{pizza}, which are not at all similar semantically. Our guess is that both \textit{giraffe} and \textit{pizza} share similar colors (yellow) and patterns (spots on the giraffe body and toppings on the pizza). JECL, on the other hand, is easily able to distinguish these objects, because the text descriptions expose their semantic differences. Surprisingly, JECL is also able to distinguish \textit{airplane} and \textit{kite}, which are not only visually similar, but are also semantically related. However, we are still able to observe some errors from JECL, such as examples of \textit{suitcase} and \textit{cellphone}, which are visually similar, assigned into the same cluster (cluster \#1) and \textit{clock} examples separated into two clusters: clocks on towers (cluster \#2) and indoors clocks (cluster \#9). \emph{To summarize, JECL appears to tolerate ambiguity better than other methods.}
\paragraph{Model Sensitivity to Parameters}
We have three hyperparameters, $\lambda$, $\beta$, and $\gamma$, to control the weighting of text and image, distribution regularizer, and alignment loss, respectively. JECL is designed for unsupervised learning scenarios, where the training data is limited and hyperparameters tuning is unachievable. In this section, we will show that JECL is robust to different hyper-parameters settings within reasonable range. Figure \ref{fig:pars} shows the model performance on various hyperparameter settings. On the weighting between text and image $\lambda$, we discover that the model is stable with $\lambda$ between 0.3 and 0.7 and performs the highest or close to the highest when $\lambda = 0.5$, which means JECL has equal weighting on text and images. This is not surprising. As we mention in Section \ref{sec:method}, JECL naturally learns from high confidence data points. In DEC, data points close to cluster centroids are those with high confidence which contribute more to the gradients. In multi-view setting, JECL learns from data points with consistent soft assignments of the text and image pairs with the aid of high confidence data points from all views. With this mechanism, tuning $\lambda$ is trivial and we will demonstrate it in missing view experiment. Then, we study the effect of hyperparameter for distribution regularizer term, $\beta$. From Figure \ref{fig:pars}, we can observe that the model performance remains stable but slowly deteriorates when $\beta$ is close to 1. The number of empty clusters decreases by 23\% when the regularizer is applied. Finally, we investigate the influence on the hyperparameter $\gamma$, which is used to adjust the text-image alignment loss. We use a wider range to test the $\gamma$, because we wonder whether a stronger image-text bond would help overall clustering performance. We can see that the overall model performance remains steady with $\gamma < 1$ and it drops dramatically with $\gamma > 1$. The reason for this deterioration is that the alignment loss dominates overall loss and causes the clustering to perform poorly. \emph{To summarize, JECL is robust when $\beta$ and $\gamma$ are smaller than 1.}

\paragraph{Sensitivity to Data Size}
We consider JECL performance with varying data size. In order to evaluate the performance of JECL on varying data size, we produce a subset of \textit{Coco-all} (labeled \textit{all-sub} in the figures and tables) with 8000 image-text pairs: 8 classes with 1000 image-text pairs each. We then sample this subset to vary the data size. We compare against the two state-of-the-art multi-view clustering methods, DMF-MVC and MLRSSC. Figure \ref{fig:size} shows that JECL's performance is robust until only 500 data points in each class remain and drops dramatically when only 100 data points in each class remain. Despite this drop in performance, \emph{JECL continues to outperform competitive models, DMF-MVC and MLRSSC on these smaller data sizes.}

\paragraph{Robustness to Missing Text}

Incomplete views are a common problem in multi-view clustering \cite{xu2015multi}; we cannot expect all images to be equipped with text descriptions. To analyze the robustness of JECL when text descriptions are missing, we remove text from a random set of images at varying rates on \textit{Coco-cross} and \textit{all-sub} dataset. We compare our method against two state-of-the-art incomplete multi-view clustering model, PIC \cite{wang2019spectral} and DAIMC \cite{hu2018doubly}. The missing view experimental results appear in Figure \ref{fig:missing}. JECL is competitive with PIC on both datasets and outperforms DAIMC by a large margin.  Figure \ref{fig:missingGradients} demonstrates that images with captions (orange dots) have larger value of $p_{ij}$ and contribute a larger gradient to the training process. Images with missing text have a smaller associated value of $p_{ij}$ because the second term in equation (\ref{eq:joint_target_distribution}) vanishes.


\section{Conclusion}
We present JECL, a method that learns representations from multi-modal image-text pairs for clustering analysis. JECL trains two parallel encoders with clustering layers, one for images and one for text, alternating between computing a proposed joint target distribution and minimizing KL divergence between the embedded data distribution to the computed joint target distribution. At the same time, JECL also learns to align the soft cluster assignments between images and text. JECL exhibits superior performance on various datasets, outperforming both single-view and state-of-the-art multi-view models. We further examine the robustness of JECL to sources of problems, including hyperparameter sensitivity, missing text and small data sizes.  Our results suggest that JECL is broadly effective for multi-modal clustering.






%
\bibliographystyle{IEEEtran}
\bibliography{reference1}

\end{document}